%% file: prl.tex
\documentclass[sigconf]{acmart}
\renewcommand\footnotetextcopyrightpermission[1]{}
\pdfoutput=1

\usepackage{multirow}
\usepackage[multiple]{footmisc}
\usepackage{makecell}
\usepackage{subcaption}
\usepackage{amsmath}  
\newcommand{\norm}[1]{\left\lVert#1\right\rVert} 
\usepackage{booktabs} 

\setcopyright{none}

\makeatletter
\def\@copyrightspace{\relax}
\makeatother

\begin{document}

\title{Attentional Road Safety Networks}

\author{Sonu Gupta}
\affiliation{%
  \institution{IIIT-Delhi}
}
\email{sonug@iiitd.ac.in}

\author{Deepak Srivatsav}
\affiliation{%
  \institution{IIIT-Delhi}
}
\email{deepak16030@iiitd.ac.in}

\author{A. V. Subramanyam}
\affiliation{%
  \institution{IIIT-Delhi}
}
\email{subramanyam@iiitd.ac.in}

\author{Ponnurangam Kumaraguru}
\affiliation{%
  \institution{IIIT-Delhi, IIIT-Hyderabad}
}
\email{pk@iiitd.ac.in, pk.guru@iiit.ac.in}
\renewcommand{\shortauthors}{S. Gupta et al.}

\begin{abstract}
Road safety mapping using satellite images is a cost-effective
but a challenging problem for smart city planning. The scarcity of labeled data, misalignment and ambiguity makes it hard for supervised deep networks to learn efficient embeddings in order to classify between safe and dangerous road segments. In this paper, we address the challenges using a region guided attention network. In our model, we extract global features from a base network and augment it with local features obtained using the region guided attention network. In addition, we perform domain adaptation for unlabeled target data. In order to bridge the gap between safe samples and dangerous samples from source and target respectively, we propose a loss function based on within and between class covariance matrices. We conduct experiments on a public dataset of London to show that the algorithm achieves significant results with the classification accuracy of 86.21\%. We obtain an increase of 4\% accuracy for NYC using domain adaptation network. Besides, we perform a user study and demonstrate that our proposed algorithm achieves 23.12\% better accuracy compared to subjective analysis.
\end{abstract}


\maketitle

\input{samplebody-conf}

\bibliographystyle{ACM-Reference-Format}
\bibliography{sample-bibliography}

\end{document}

%% file: samplebody-conf.tex
\section{Introduction}
\label{sec1}
Road accidents remain one of the pressing communal welfare concerns. Regardless of notable advancements in the field of vehicle technology and road engineering, on a global scale, traffic accidents are one of the leading causes of premature death and injury. According to official statistics from the World Health Organization (WHO), more than 1.25 million people die every year due to traffic accidents. Besides, traffic accidents cost many countries up to 3\% of their GDP \citep{world2015global}. Therefore, minimizing the road accidents is a worldwide challenge and can benefit a majority of the nations in different ways. Towards this, it is important to understand which road segments are potentially dangerous or safe. 

Few works have shown the influence of environmental factors like weather, light condition on road accidents \citep{tamerius2016precipitation,chung2018framework,black2018effects}. 
However, gathering such data is costly and laborious. Further, due to lack of resources and technology, such data is not maintained properly in most low and middle-income nations and unfortunately, these are the nations which suffer dreadfully from traffic accidents \citep{schmucker2010road}. Hence, there is a need for an efficient approach which can work well with easily available and affordable data. To this end, satellite images are used for road safety mapping \citep{najjar2017combining}. However, the dataset is mostly imbalanced, and percentage of the safe class is far more compared to the dangerous class. Further, the images are misaligned. Thus, using such data for training efficient supervised models is another challenge. 

\begin{figure} [htbp]
	\centering
	\includegraphics[width=0.46\textwidth]{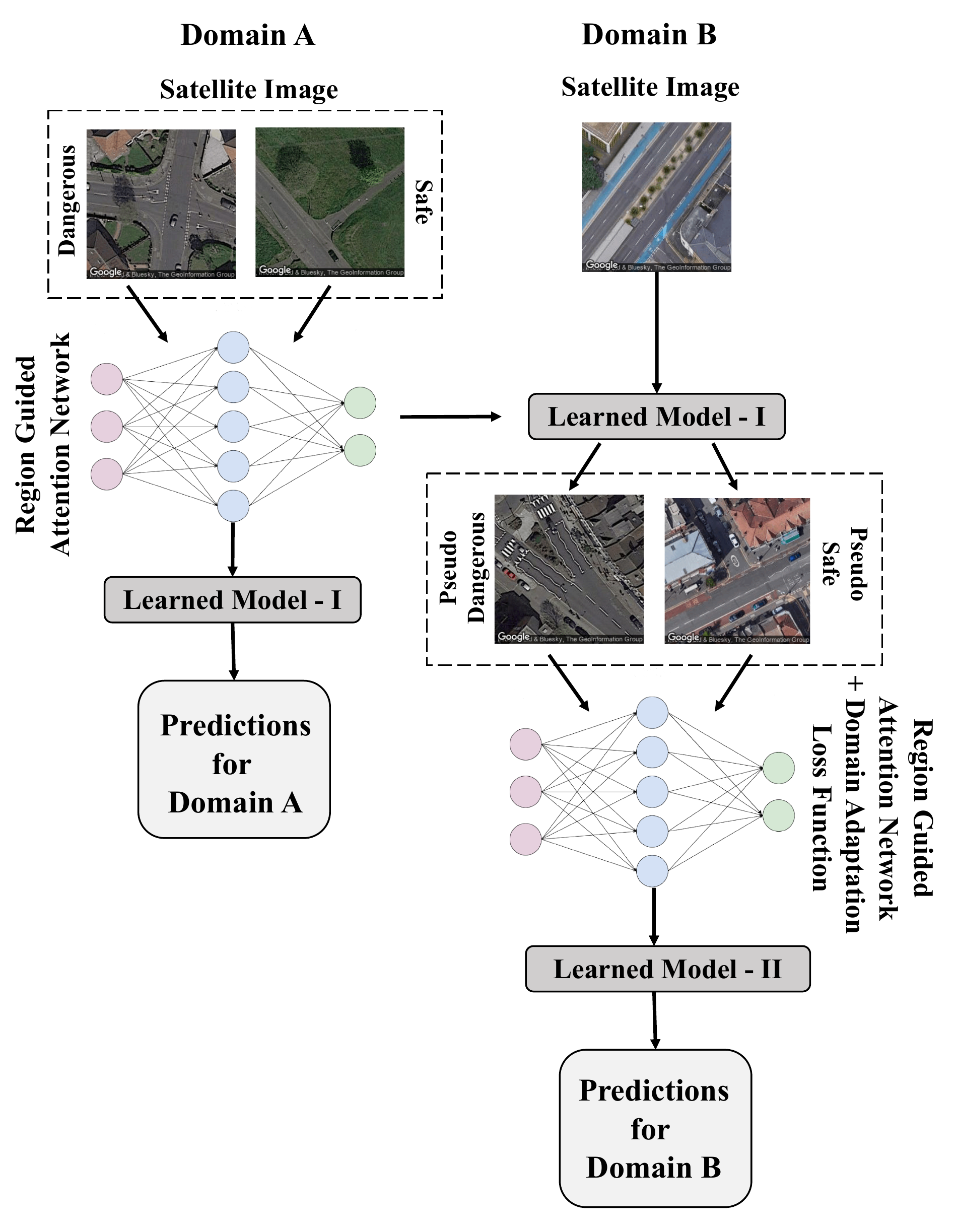}
    \caption{A deep learning framework to predict city-scale safety maps using raw satellite images for Domain A (Source city) and Domain B (Target city) such that labels are available only for Domain A.}
    \label{fig:teaser_diagram}
\end{figure}

In addition, it has been observed that models trained using classical machine learning techniques for one region do not perform well if tested on regions that differ immensely in terms of traffic regulations, city planning, architecture, etc. \citep{najjar2017combining}. Also, it is quite inconvenient to obtain traffic accident data for every different region and train a model for the same. 
Therefore, the difference in domains is yet another challenge in the effective road safety mapping.

To address these challenges, we propose a deep region guided attention network as shown in Figure \ref{fig:teaser_diagram}. We use a base network (ResNet-50) to extract global features. We further use a sub-network which can attend to individual subregions. Towards this, we extract the conv 2 layer features and divide them into $N$ non-overlapping regions. Region or part based networks have shown good accuracy in re-identification tasks \citep{zhu2017part}. We further augment this network to adapt to different domains. We use a loss function based on the covariance matrix to minimize the gap between the source and target domains. Our contributions are:

\begin{itemize}
  \item We propose a deep learning framework that uses the region guided attention network to predict accurate city-scale safety maps from satellite images.
  \item We propose a domain adaptation network with a training loss which minimizes the Frobenius norm of the difference of \textit{within} class covariance matrices of source and target, as well as the difference of \textit{between} class covariance matrices of source and target.
 \end{itemize}

The rest of this paper is organized as follows. The related work is discussed in section \ref{rw}. Section \ref{overview} presents the data sources and collection technique and also introduces the formulation of the problem. Our approach is explained in section \ref{dan}. Section \ref{us} shows a user-study followed by Section \ref{ex}, which describes our experiments. Finally, the paper is concluded in section \ref{conc}.

\section{Related Work}
\label{rw}
In this section, we briefly outline an overview of studies on accident hotspot detection (also known as, black spots and black zones), understanding road accidents, and lastly, the city-scale road-safety mapping that exploits recent advancements in the field of deep learning.

There are few studies on the detection of high-density accident hotspots. Kernel density estimation and clustering have been used to create classifiers to identify accident hotspots \citep{anderson2009kernel,bil2013identification}. Another interesting work includes the use of Bayesian statistics to predict accidents \citep{deublein2013prediction}. \citeauthor{fawcett2017novel} presented a Bayesian hierarchical model to rank accident hotspots in line with their possibility to surpass a threshold accident count in some future time period. Researchers have developed models to estimate the total number of crashes, the number of injury crashes, and the number of property damages and proved that the models are statistically meaningful and closer to real-world data \citep{pulugurtha2013traffic}. In addition, association rules have been used for the identification of accident situations that frequently occur together \citep{geurts2003profiling,geurts2005understanding,das2018factors}. \citeauthor{gomes2013influence} proposed a model for accident frequency estimation which takes the influence of the road characteristics into account.

\begin{table*}
    \caption{Sample traffic accident reports for London.}
    \label{t:tab1}
    \centering
  \begin{tabular}{|c|c|c|c|c|c|c|c|}
    \hline
    ID & Date & Time & Day of the week & Latitude & Longitude & No. of Vehicles & No. of Casualties\\
    \hline
1 & 01/11/2016 & 2:30:00 AM & 3 & 51.5847 & 0.2793 & 2 & 1 \\ \hline
2 & 21/11/2016 & 6:00:00 PM & 2 & 51.5092 & 0.0472 & 2 & 2 \\ \hline
3 & 20/05/2016 & 7:00:00 PM & 6 & 53.8126 & -2.9323 & 1 & 1 \\ \hline
4 & 11/01/2016 & 7:07:00 AM & 2 & 54.9785 & -1.6203 & 2 & 3 \\ 
    \hline
    \end{tabular}
\end{table*}

In the recent past, the remarkable progress in deep learning has contributed significantly to the field of computer vision. Spatio-temporal data have been used by researchers to predict the number of accidents in a given area with the help of ConvLSTM \citep{yuan2018hetero}. \citeauthor{chen2016learning} developed a Stacked Denoising Autoencoder for prediction of traffic accident risk level at the city-scale using real-time GPS data of users. In a similar study, \citeauthor{najjar2017combining} demonstrate that visual attributes captured in satellite image can be used as a proxy signal of road safety. They proposed a deep-learning based mapping framework that exploits open data to predict city-scale safety maps at bearable costs. However, the distinction among the above frameworks resides in their application itself. Both \citep{chen2016learning,yuan2018hetero} are interested in the real-time prediction of traffic accidents, whereas \citep{najjar2017combining} is inclined towards assisting in informed decision-making for city-planning and policy formulation when heterogeneous data is not accessible or bearable. Our work is closer in spirit to \citeauthor{najjar2017combining}; however, there are three major differences. First, we only consider safe and dangerous classes compared to an additional neutral class defined by \citeauthor{najjar2017combining} This is because a neutral road segment may not assert a strong decision making during road safety planning. Second, 
the authors extract features from AlexNet, whereas, we propose a novel architecture. Third, we also perform domain adaptation for the unlabeled target domain. Our two major goals are as follows. First, we propose
a region guided attention network to generate a city-scale safety map. Second, we propose a domain adaptation technique to generate safety maps where traffic accident data is not available.

Domain Adaptation is a branch of machine learning in which we aim at learning a well performing network from a source data that can generalize well on a different but related target data distribution \citep{ben2010theory}. It has been widely studied in visual applications \citep{csurka2017domain,wu2018dcan}. One of the popular class of algorithms is discrepancy based methods \citep{tzeng2014deep,long2015learning,sun2016deep}. In particular, \citeauthor{sun2016deep} propose to minimize the Frobenius norm of the difference between the feature covariance matrix of source and target. In our case, we also make use of the labels available for source and pseudo-labels for the target. These pseudo-labels are determined by the model trained on the source data. We then compute the feature covariance matrix for safe and dangerous classes respectively. Further, we compute our domain adaptation loss function based on \textit{within} and \textit{between} class covariance matrices of source and target.

\section{Overview}
\label{overview}
In this section, we present the data that we use at various stages of the work. We also introduce the formulation of our problem.
\subsection{Data}
\begin{figure}[htbp]
\centering
\subcaptionbox{London\label{sfig:london}}{\includegraphics[width=2.7cm,height=2.7cm]{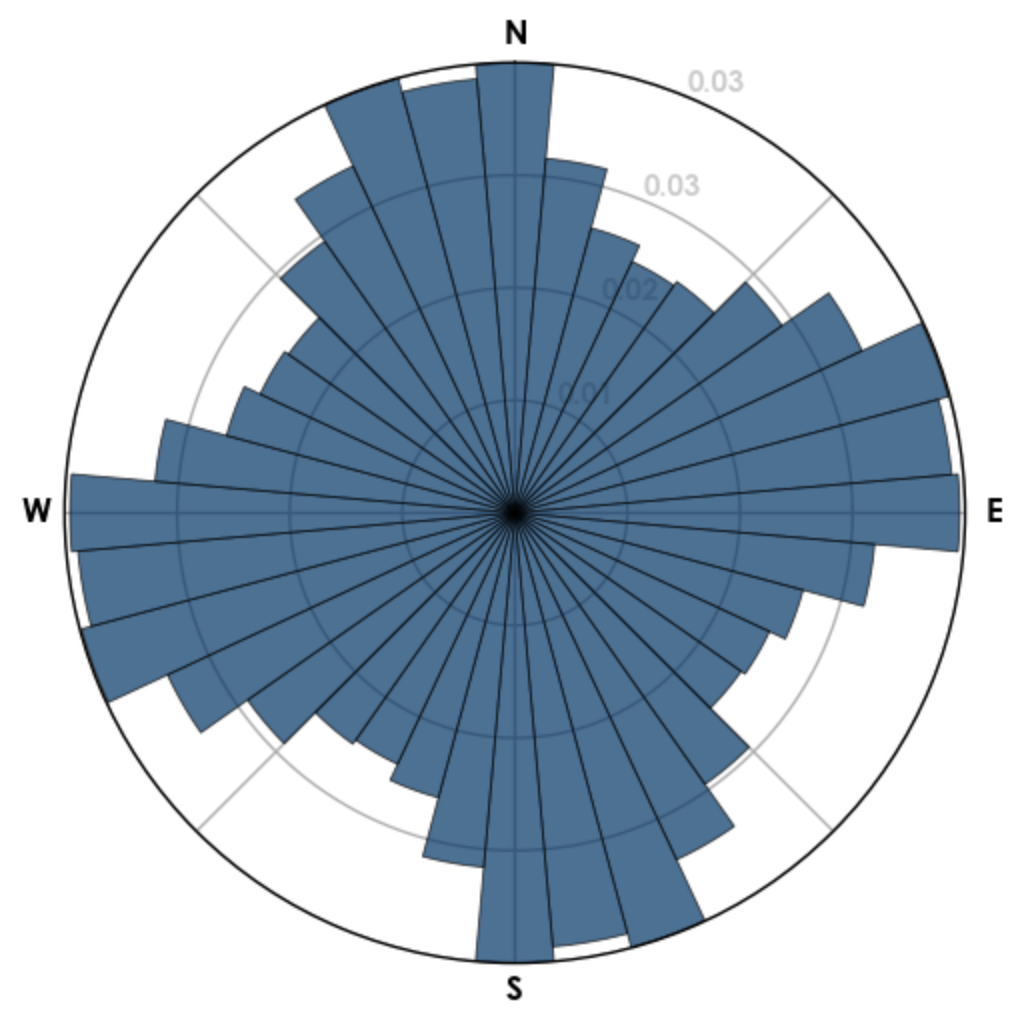}}\hfill
\subcaptionbox{Manhattan, NYC\label{sfig:manhattan}}{\includegraphics[width=2.7cm,height=2.7cm]{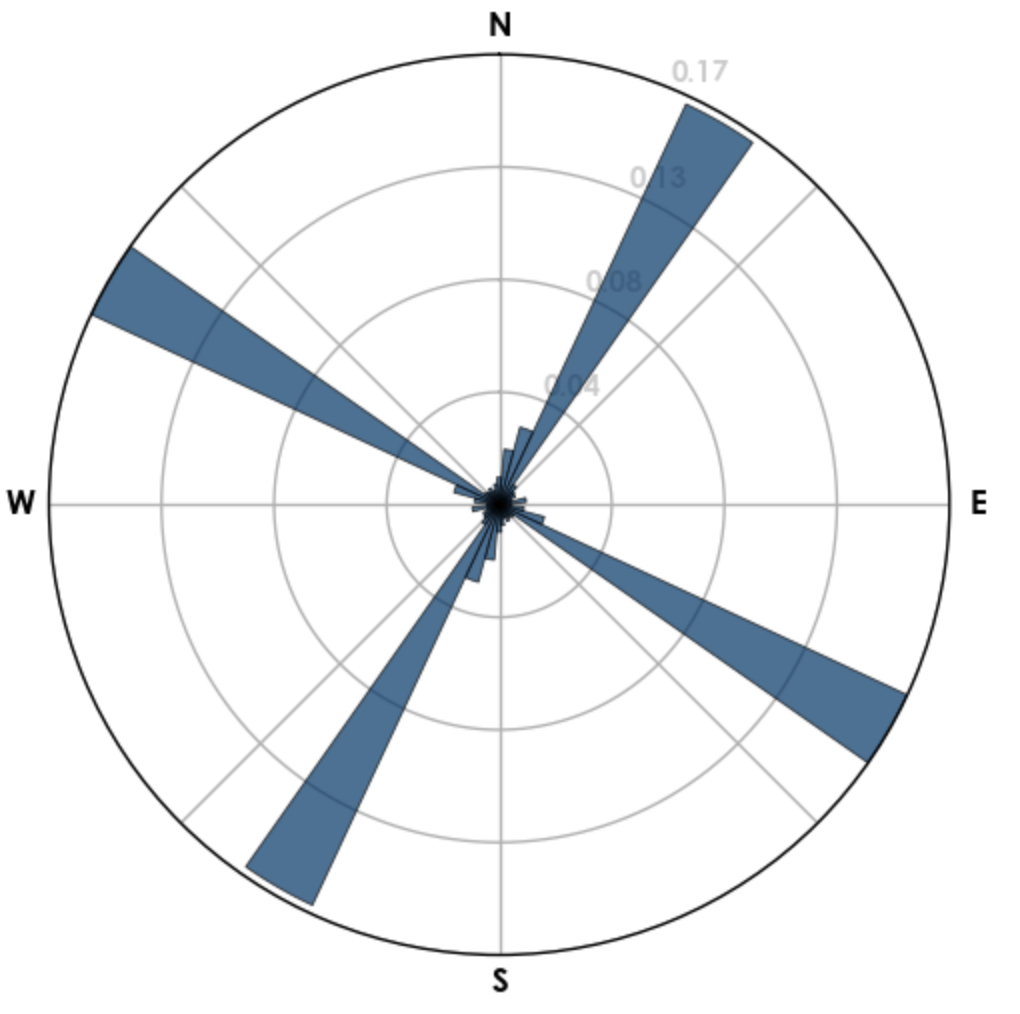}}\hfill
\subcaptionbox{Denver\label{sfig:denver}}{\includegraphics[width=2.7cm,height=2.7cm]{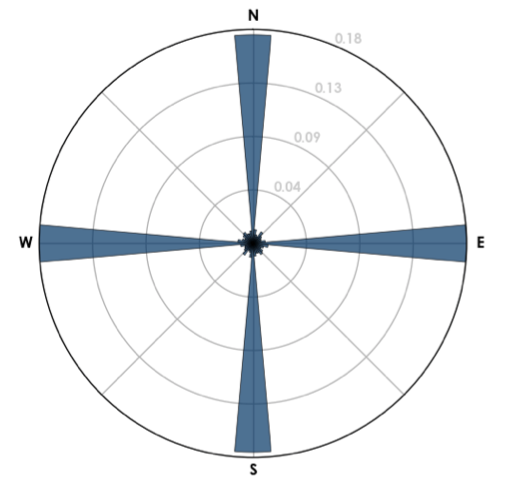}}\\
\caption{The spatial differences in terms of street network orientation. All the locations have visual difference in their infrastructure orientation which makes learning a single network for all different domains a challenging task.}
\label{fig:spatial_difference}
\end{figure}

We select two cities, London (UK) and New York City (USA) for the purpose of this study. Our reason for selecting these two cities is two-fold. First, the availability of sufficient data to work on and second, the extreme difference in traffic regulations, city planning, architecture etc. Figure \ref{fig:spatial_difference} shows the difference in the street network orientation of both the cities \citep{boeing2018urban}. All the data that we use in this work are available as \textit{open data}. Open data is the scheme under which some datasets are freely available to use and republish, without any restrictions \citep{dietrich2009open}. We also present results for Denver (USA) which is altogether a different domain. 

\textbf{Traffic Accident Data:} We collected 99,516 traffic accident records for London from 2013 to 2016.\footnote{https://www.europeandataportal.eu/data/en/dataset/road-accidents-safety-data} We gathered 1,256,205 traffic accident reports for New York City (NYC) from 2012 to 2017.\footnote{https://opendata.cityofnewyork.us} Similarly, we collected 143,776 traffic accident reports for Denver from 2013 to 2017.\footnote{https://www.denvergov.org/opendata/dataset/city-and-county-of-denver-traffic-accidents} Each record has numerous attributes such as latitude, longitude, date, time, vehicle type etc. However, there is a difference in the datasets. Table \ref{t:tab1} presents sample records from London data.
Information such as junction type, weather etc. is only available in London dataset. Therefore, we restrict ourselves to common attributes only.

\textbf{Satellite Images:} We collect satellite images using Google's Static Maps API.\footnote{https://cloud.google.com/maps-platform/}

\subsection{Problem Formulation}
We divide the whole study area by imposing a grid \textit{G}, where each grid $g_i$ is a square region of $s \times s$. Here, $s$ is equal to $30$ meters which is a sufficient area for traffic accident analysis \citep{najjar2017combining}. In this work, our objective is to produce a model that learns features such that it can predict the safety level of a given grid $g_i$ using its raw satellite image.

In order to curate a training dataset, we plot each accident on the grid using the locations from the Traffic Accident data. Each grid $g_i$ is given a safety score $S_i$ which is equal to the number of accidents in the grid. The higher $S_i$ corresponds to higher number of accidents.
 
We obtain satellite image for each of these grids. To obtain the image label for each $g_i$, we apply the classical k-means algorithm to bin $S_i$ into two clusters. Hence, we get two clusters of \textit{safe} and \textit{dangerous} locations. We observe that the obtained set of labels are highly imbalanced. Over 88\% of the regions are labeled as safe. Therefore, we re-sample the dataset by down-sampling the majority class such that both classes are balanced out.

\begin{figure*}[ht]
	\centering
	\includegraphics[width=\textwidth, height = 13cm]{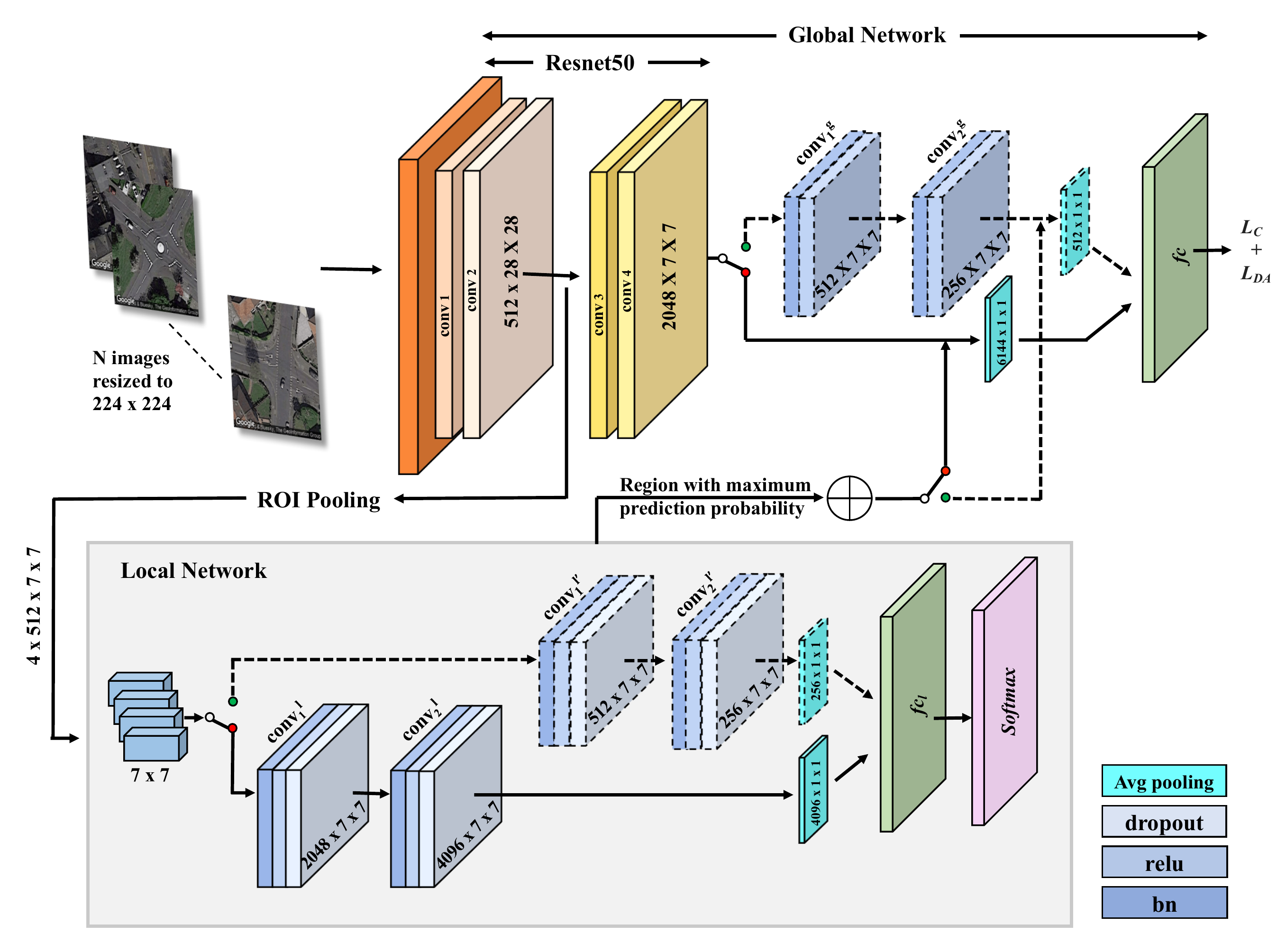}
    \caption{Architecture of our proposed network. There are three 2-way switches that enable the network to run in two modes. When all the switches are connected to \textit{red} nodes, DAM mode is enabled whereas if switches are connected to green nodes, DAM-DA mode is enabled. The solid and dashed lines represent DAM and DAM-DA mode respectively. It consists of a global and a local network. ResNet50 acts as a base to the global network. The ROI pooled features from conv 2 layer are passed through the local network to identify the subregion with maximum prediction probability to guide the global network. $L_C$ is used for training DAM and is also used along with $L_{DA}$ for training DAM-DA (Best viewed in color).}
    \label{fig:arch_dia}
    
\end{figure*}

\section{Deep Attention-Network Approach}
\label{dan}
In this section, we begin by introducing the network architecture. Then we elaborate 
the technique used to implement the unsupervised domain adaptation for satellite images that belong to different domains. 

\subsection{Network Architecture} \label{na}
The model consists of a global and a local network, with the global network borrowing layers from ResNet50 \citep{he2016deep}. We show the architecture of the proposed model in Figure \ref{fig:arch_dia}.
There are three 2-way switches in the architecture which enables the network to run in two different modes. When all the switches are connected to \textit{red} nodes, the Deep Attention Model (DAM) mode is enabled whereas if switches are connected to all the green nodes, the Deep Attention Model with Domain Adaptation (DAM-DA) mode is enabled. The solid and dashed lines represent DAM and DAM-DA mode respectively. We discuss the DAM-DA mode in the next section.
As shown in Figure \ref{fig:arch_dia}, the global network is augmented with a local network. In our experiments, we notice that the global network based on ResNet only, gives significant false positives, where false positives are defined as dangerous regions being predicted as safe. A possible reason behind such false positives could be that the satellite images are usually misaligned. To overcome misalignment and improve the performance of ResNet, we design the network such that it can exploit local regions in addition to the global region. We perform \textit{Region of Interest} (ROI) pooling on conv 2 layer of the model with the subregions being $N$ non-overlapping blocks. We pass features of each subregion through a local network. The local network consists of conv$_{1}^{l}$ and conv$_{2}^{l}$ layers. Each subregion of size 7x7 acts as an input to the local network to extract features and to identify which of them is useful for the classification task. Thus, this network attends to the regions which can significantly contribute towards enhancing the classification accuracy. We perform average pooling on the output of conv$_{2}^{l}$ layer. We obtain a feature vector of size 4096x1x1 that is passed through $fc_l$. The subregion with the maximum prediction probability of either of the two classes (Safe and Dangerous) is concatenated to the feature-map of the global network. We perform concatenation after the conv 4 layer of the global network. \textit{Thus, we can see that only certain subregions can guide the global feature vector for classification.} Once average pooling is performed, we pass the feature vector through $fc$ to obtain the final $d$-dimensional vector. We train the network using cross entropy loss $(L_C)$.

\subsection{Domain Adaptation}
In this section, we discuss our proposed Domain Adaptation (DA) technique for the target domain where there is no annotated data available. In our experiments, we consider London as the source domain for which labeled data is present, whereas, NYC as the target domain for which labels are not given. There are significant differences in both the cities. 
Although NYC is nearly 20\% smaller in terms of area, it's population density is almost double than that of London. 
According to \citeauthor{nieminen2002population}, higher the population density, higher are the chances of occurrence of accidents. Due to above-mentioned reasons, these cities would be a good extreme example for performing DA and can give us an idea of lower bound of model's performance.

\begin{table}[ht]
\caption{The comparison of the London and NYC in terms of area and population density.}
\label{t:differences_in_cities}
\centering
\begin{tabular}{|c|c|c|}
\hline
 & London & NYC \\ \hline
\begin{tabular}[c]{@{}c@{}}Area \footnotesize{(mi\textsuperscript{2})} \end{tabular} &607 & 468.5 \\ \hline
\begin{tabular}[c]{@{}c@{}}Population Density\\ \footnotesize{(people per mi\textsuperscript{2})} \end{tabular} &14,550 & 28,491 \\ \hline
\end{tabular}
\end{table}

First, we train DAM network using data from the source domain as explained in the section \ref{na}. Using this trained model, we generate \textit{pseudo labels} for data from the target domain. Thus, we have labels for the source domain (London) and pseudo labels for the target domain (NYC). Now, we use data from both source and target domain to train the augmented DA (DAM-DA) network as shown in Figure \ref{fig:arch_dia}. There are three differences when compared to DAM. 
First, this network has two additional convolutional layers in the global network, conv$_{1}^{g}$ and conv$_{2}^{g}$ layers of dimensions 512x7x7 and 256x7x7 respectively, to reduce the dimensions of the feature maps. This augmentation is necessary to be able to compute the covariance matrix efficiently. We explain the utility of the covariance matrix in the later part of this section. Second, instead of conv$_{1}^{l}$ and conv$_{2}^{l}$ layers in the local network, we use conv$_{1}^{l'}$ and conv$_{2}^{l'}$ layer of size 512x7x7 and 256x7x7 respectively. Again, this is done to reduce the dimensions for covariance matrix calculation. Third, in DAM mode, we train the network using $L_C$ only, but in DAM-DA mode we use domain adaptation loss ($L_{DA}$) along with $L_C$. We explain $L_{DA}$ next. 

Let $x_i$ denote the feature for \textit{i}-th sample classified as dangerous in source domain. Similarly, let $y_i$ be the feature for \textit{i}-th sample classified as safe in source domain. These features are obtained from the final layer $(fc)$ of the network as shown in Figure \ref{fig:arch_dia}. Now, we can obtain the \textit{within} class covariance matrix $\sum_{SW}$ $\in \mathcal{R}^{d\times d}$ as,

\begin{equation}
\begin{split}
\sum\nolimits_{SW} = \sum\nolimits_{i,j \neq i} (x_{i}-x_{j}) (x_{i}-x_{j})^T  \\ + \sum\nolimits_{i,j \neq i} (y_{i}-y_{j}) (y_{i}-y_{j})^T
\end{split}
\end{equation}

Similarly, we can compute the \textit{between} class covariance matrix $\sum_{SB}$ $\in \mathcal{R}^{d\times d}$ as,
\begin{equation}
\sum\nolimits_{SB} = \sum\nolimits_{i,j}(x_{i}-y_{j}) (x_{i}-y_{j})^T
\end{equation}

We can compute the \textit{within} and \textit{between} class covariance matrices for target domain in a similar manner. Let these be denoted by $\sum_{TW}$ and $\sum_{TB}$ respectively. Then, we use the following loss function to adapt to the target domain,

\begin{equation} \label{eq:loss}
L_{DA} = \norm{ \sum\nolimits_{SW} - \sum\nolimits_{TW} }_F^2 +  \norm{\sum\nolimits_{SB} - \sum\nolimits_{TB} }_F^2
\end{equation}
where $\norm{.}_F$ denotes the Frobenius norm. Though the pseudo labels are noisy, it is still beneficial to use them. In our experiments, we demonstrate that 
using the loss to minimize the distance between feature covariance matrices between source and target, which is agnostic of source labels, achieves sub-par performance when compared to $L_{DA}$.

\section{User Study}
\label{us}
We conduct a study with an aim to observe that how accurately humans can classify raw satellite images as safe or dangerous. It's first of its kind user study. We develop an annotation portal and provide the raw satellite images from our test-data to the users. 
The portal has two sections. In the first section, we provide four sample images from each category. In the second section, an image is presented to the user along with three options to choose from. A user can mark an image as \textit{dangerous, safe or unsure}. We show the same image to three different users and record their response to maintain the confidence in the user's response. We also add sixteen more sample images from each class to help the user. We record users' response for 1,000 images having 500 images from each category, i.e., safe and dangerous classes. 

In this study, 38 users participated. All the participants are above 18 years of age. To check inter-annotator agreement, we compute Fleiss' kappa \citep{fleiss1973equivalence}. We achieve 0.2743 which depicts a fair agreement. To assign a final label to the image, we calculate the mode of all three responses. If all the responses are different, we discard that response. In our study, we found 0.03\% of such responses and discarded them. Finally, 67.22\%, 27.94\%, and 4.84\% images are marked as \textit{safe, dangerous,} and \textit{unsure} respectively by users. In Table \ref{t:user_study_data}, we provide a complete description of the user-study results. With 80.32\% accuracy, users are able to detect \textit{safe} images whereas the accuracy drops to 45.02\% for \textit{dangerous} images. Therefore, in this case, we find that humans can identify safe locations with better accuracy than dangerous locations.

\begin{table}[htbp]
\caption{The complete description of the user-study outcome (in \%).}
\label{t:user_study_data}
\centering
\begin{tabular}{c|c|c|c|c|}
\multicolumn{2}{c}{}
            &   \multicolumn{3}{c}{Predicted} \\
            \cline{2-5}
    &       &   Safe &   Dangerous     & Unsure       \\ 
    \cline{2-5}
\multirow{2}{*}{\rotatebox[origin=c]{90}{Actual}}
    & Safe   & 40.41   & 05.56      &  4.32        \\ \cline{2-5}
    & Dangerous    & 26.80    & 22.37   & 0.51            \\ 
    \cline{2-5}
    \end{tabular}
\end{table}

\section{Experiments}
\label{ex}
\begin{figure}[htbp]
\centering
\subcaptionbox{Dangerous location, might be perceived as \textit{safe.}\label{sfig:ds-1}}{\includegraphics[width=3.5cm,height=3.5cm]{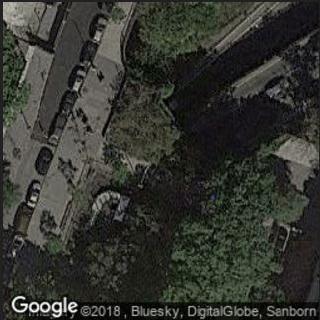}} \hspace{4mm} 
\subcaptionbox{Safe location, might be perceived as \textit{dangerous.}\label{sfig:sd-1}} {\includegraphics[width=3.5cm,height=3.5cm]{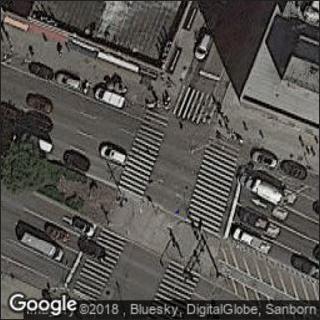}}\hfill
\caption{Image Samples to show the non-triviality of learning from satellite data.}
\label{fig:non_trivial}
\end{figure}

\begin{figure*}[ht]
\centering
\rotatebox{90}{\hspace{0.7cm}Dangerous Image}
{\includegraphics[width=4cm,height=4cm]{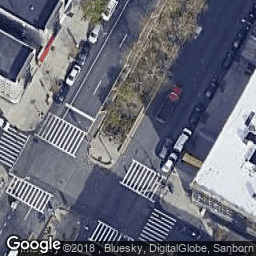}}\hspace{4mm}
{\includegraphics[width=4cm,height=4cm]{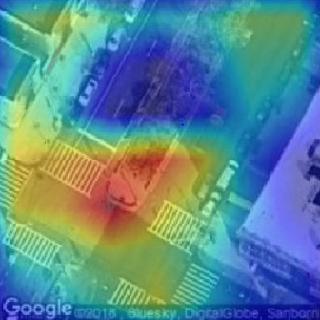}} \hspace{3mm}
{\includegraphics[width=4cm,height=4cm]{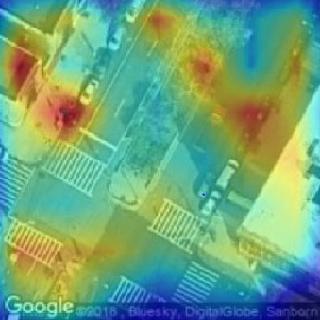}} \hspace{3mm}
{\includegraphics[width=4cm,height=4cm]{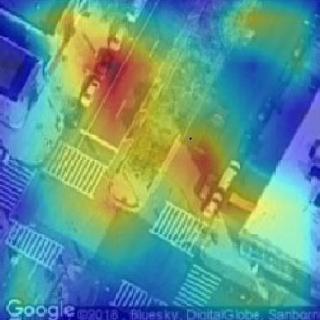}}\\
\vspace{3mm}
\rotatebox{90}{\hspace{12mm}Safe Image}
\subcaptionbox{Original Images\label{sfig:Safe-incorrect-original}}
{\includegraphics[width=4cm,height=4cm]{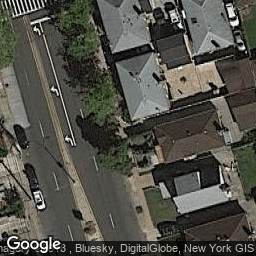}}\hspace{4mm}
\subcaptionbox{ResNet50\label{sfig:CAM-Resnet2}}
{\includegraphics[width=4cm,height=4cm]{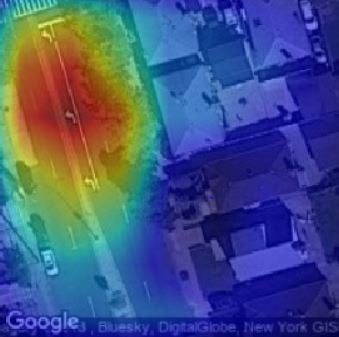}}\hspace{4mm}
\subcaptionbox{DAM-Global Network\label{sfig:CAM-DAM-11}}{\includegraphics[width=4cm,height=4cm]{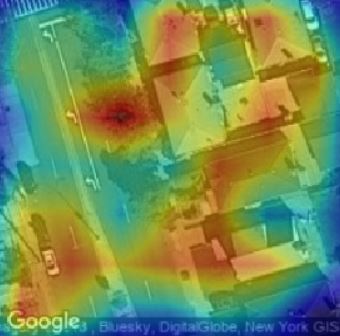}}\hspace{4mm}
\subcaptionbox{DAM-local Network\label{sfig:CAM-DAM-21}}{\includegraphics[width=4cm,height=4cm]{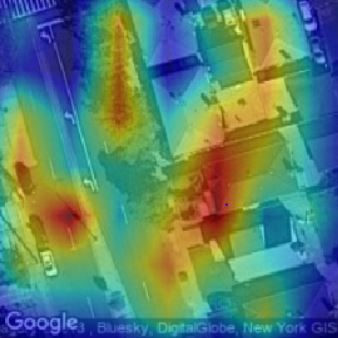}}\\
\caption{A comparison of activated features of a \textit{dangerous} and a \textit{safe} location among ResNet50, DAM-Global Network, and DAM-Local Network. Top row and bottom row represent dangerous and safe locations respectively. DAM-Global Network contains activated features in the global network before the concatenation of local network features. DAM-Local Network contains activated features from the local network alone (Best viewed in color).}
\label{f:cams}
\end{figure*}

We initially performed the experiments with AlexNet, VggNet, DenseNet, and ResNet and found that ResNet gives the best accuracy.
We now prove the efficacy of the proposed technique by assessing the classification accuracy of the predicted road safety maps in multiple scenarios. It is also important to note the non-triviality of learning from satellite data, as it is seen in the user study, and can also be seen in the examples shown in Figure \ref{fig:non_trivial} and Figure \ref{f:cams}.

\textbf{Training Parameters:} We train our network on London dataset. There are 4,517 training samples and 903 validation samples in each class. The models are trained with a batch size of 4, with learning rate as 0.0001 and a learning rate decay of 0.5 per 10 epochs. We train it over 50 epochs on a Nvidia GeForce TITAN X GPU.

\textbf{Evaluation of DAM:}
To assess the model, we test on unseen data from London. As shown in Table \ref{t:tab2}, we test the models in multiple scenarios such as with horizontal (DAM + HS), vertical (DAM + VS), and square boxed (DAM + SQ) subregions and with the combinations of them. We test another variation of the model where the activation maps are produced based on an intermediate convolutional layer. We experiment with two conv layers; conv 2 and conv 3 layer. We finally use conv 2 layer as we achieve better results with it. We find that the DAM using HS+VS+SQ subregions outperformed ResNet50, VGG19, and its other variants on the same dataset as shown in Table \ref{t:tab2}.
\begin{table}[htbp]
\caption{Comparison of classification accuracy (in \%) for ResNet50, VGG19, and variants of the DAM both for original dataset (London) and cross dataset (NYC and Denver).}
\label{t:tab2}
\centering
\begin{tabular}{|c|c|c|c|}
\hline
Model& Original Data & \multicolumn{2}{c|}{Cross Data} \\ \cline{2-4}
 & London & NYC & Denver \\ \hline
ResNet50 & 85.77  & 69.16  & 70.00  \\ \hline
VGG19 & 85.83  & 64.60  & 70.00  \\ \hline
DAM (HS) & 85.81  & 72.28  & \textbf{76.20}  \\  \hline
DAM (VS) & 85.52  & \textbf{74.77}  & 75.00  \\ \hline
DAM (SQ) & 85.86  & 70.70  & 70.00  \\ \hline
DAM (HS+VS) & 85.34  & 70.37  & 70.01  \\ \hline
DAM (HS+VS+SQ) & \textbf{86.21}  & 67.23  & 69.86 \\  \hline
\end{tabular}
\end{table}

\textit{One of our goals is to minimize the false positives.} This is necessary as the cost of predicting safe as dangerous may only lead to overly conservative approaches for planning, whereas, having high false positive rate with more dangerous locations being marked safe can lead to fatal accidents. Therefore, we consider the classification of dangerous images as safe to be more costly than vice-versa. We test ResNet50 and DAM on London, NYC, and Denver test-set of 7,228, 8,342, and 500 images respectively. From Table \ref{t:tab4}, we can see that DAM gives fewer false positives in comparison to ResNet50 for every domain.
\begin{table}[htbp]
\centering
\caption{A comparison of false positive rate between DAM and ResNet50. All results are in percentage.}
\label{t:tab4}
\begin{tabular}{|c|c|c|c|}
\hline
Model & London & NYC & Denver \\ \hline
DAM & \textbf{07.60}  & \textbf{22.05}  & \textbf{29.20} \\ \hline
ResNet50 & 12.73  & 39.05  & 40.80  \\ \hline
\end{tabular}
\end{table}

\textbf{Cross Dataset Testing:}
We also perform cross-data testing and test our model on NYC and Denver dataset. From Table \ref{t:tab2}, we can see that DAM + VS and DAM + HS performs the best for NYC and Denver dataset respectively. For DAM (HS+VS+SQ), the base model trained on London with three sub-regions is over-fitting, and the network does not generalize well on Denver. When we decrease sub-regions to 2, the accuracy increases for NYC and Denver, and the model performs the best when used with only one sub-region. 

\textbf{Qualitative Analysis:}
In order to understand the behavior of the network qualitatively, we generate Class Activation Maps (CAM) \citep{zhou2016learning}. In Figure \ref{f:cams}, the images in the top row correspond to a dangerous image and its CAMs for ResNet50, DAM-Global Network, and DAM-Local Network. Similarly, the images in the bottom row correspond to a safe image and its corresponding CAMs. From the top row in Figure \ref{sfig:CAM-DAM-11} and Figure \ref{sfig:CAM-DAM-21}, we can establish that the DAM not only identifies the green region around the divider on the road but also identifies the cars and roads in the surrounding areas as well, which positively contributes towards the identification of this image as dangerous. This image contains a dangerous location, but, ResNet50 mis-classifies it as safe. As from top row in Figure \ref{sfig:CAM-Resnet2}, ResNet50 seems to have identified the green region and used that to classify the sample as safe, whereas our proposed DAM is efficient enough to correctly identify it as a dangerous location. Similarly, from the CAM in the bottom row, we observe that ResNet50 recognizes only roads whereas the DAM identifies houses and trees in the image. Therefore, DAM correctly identifies this sample as safe whereas ResNet50 mis-classifies it as dangerous.

\textbf{Evaluation of DAM-DA Network:}
We train our DAM-DA network with 2,085 images of each class, i.e., Safe and Dangerous. We use 4,170 images each from the source (London) and target (NYC) domain. We use a batch size of 16, a learning rate of 0.0001 and a decay of 0.5 per 10 epochs. It is trained for over 50 epochs. We test the DAM-DA network on 3,336 images from NYC. As shown in Table \ref{t:tab6}, 
DAM-DA network gives an accuracy of 75.75\% in comparison to DAM network which gives an accuracy of 71.94\%. 

In another setting, we replace $L_{DA}$ with the loss \citep{sun2016deep} in Equation \ref{lda},

\begin{equation} \label{lda}
L_{DA}^{ST} = \frac{1}{d}\norm{C_{S} - C_{T} }_F^2 
\end{equation}

where $C_{S}$ and $C_{T}$ denotes the feature covariance matrix of source and target samples in a training batch, respectively. When compared to $L_{DA}$, $L_{DA}^{ST}$ does not take the class labels into account. As shown in Table \ref{t:tab6}, we obtain an accuracy of 74.73\%. Similarly, DAM$-$DA$-L_{DA}$ results in the least false positives rate (18.94\%) in comparison to its counterparts. Thus, we can see that the class wise covariance loss works better in this scenario. 

\begin{table}[htbp]
\caption{A comparison of DAM, DAM-DA with $L_{DA}$, and DAM-DA with $L_{DA}^{ST}$ network with batch-size of 16.}
\label{t:tab6}
\centering
\begin{tabular}{|c|c|c|}
\hline
Model & Accuracy (\%) & FPR(\%)\\ \hline
DAM & 71.94 & 30.03\\ \hline
DAM$-$DA$-L_{DA}$ & \textbf{75.75} & \textbf{18.94} \\ \hline
DAM$-$DA$-L_{DA}^{ST}$ & 74.73 & 30.69\\ \hline
\end{tabular}
\end{table}

\textbf{Comparison of DAM and Humans:} We test DAM on the set of same 1000 images that we use for user study. We present these results in Table \ref{t:user_study_comp}.
We can see that DAM is more accurate than humans by a significant difference of 23.12\%. Moreover, there is a substantial difference of 30.77\% in precision. Therefore, this proves that DAM performs better than humans in this case. 

\begin{table}[htbp]
\caption{A comparison of the performance of our proposed deep learning model (DAM) and humans.}
\label{t:user_study_comp}
\centering
\begin{tabular}{|c|c|c|c|}
\hline
Model & Acc(\%) & Precision & F1\\ \hline
Humans & 62.78  & 0.601 & 0.687\\  \hline
DAM & \textbf{85.90}  & \textbf{0.909} & \textbf{0.850}\\ \hline
\end{tabular}
\end{table}

\section{Conclusion}
\label{conc}
In this paper, we address the challenge of learning efficient embeddings to classify the road segments as dangerous or safe using easily available and inexpensive data. We leverage open data and satellite images to predict city-scale road safety maps. We propose a deep learning based model that uses a region guided attention network. It consists of a global and a local network. The local network attends to features in the subregions and the features with maximum prediction score are used to guide the global features to enhance the accuracy. We evaluate our network on the public dataset of London and achieve the accuracy of 86.21\%. We experiment with the cross-datasets of NYC and Denver and achieve significant results with the accuracy 74.77\% and 76.20\% respectively. In addition, we propose a covariance loss based domain adaptation for the scenario where target domain labels are missing. In our experiments, we show that with the domain adaptation network, the accuracy of NYC increases by 4\% and the network also achieves the lowest false positives. We also conduct a user study and find that our model outperforms human by 23.12\%. In the future, we would like to explore Places trained models instead of ImageNet trained models.

\section{Acknowledgments}
We would like to thank Precog at IIIT-Delhi for supporting this research work.

%% file: prl.bbl

\begin{thebibliography}{30}


\ifx \showCODEN    \undefined \def \showCODEN     #1{\unskip}     \fi
\ifx \showDOI      \undefined \def \showDOI       #1{#1}\fi
\ifx \showISBNx    \undefined \def \showISBNx     #1{\unskip}     \fi
\ifx \showISBNxiii \undefined \def \showISBNxiii  #1{\unskip}     \fi
\ifx \showISSN     \undefined \def \showISSN      #1{\unskip}     \fi
\ifx \showLCCN     \undefined \def \showLCCN      #1{\unskip}     \fi
\ifx \shownote     \undefined \def \shownote      #1{#1}          \fi
\ifx \showarticletitle \undefined \def \showarticletitle #1{#1}   \fi
\ifx \showURL      \undefined \def \showURL       {\relax}        \fi
\providecommand\bibfield[2]{#2}
\providecommand\bibinfo[2]{#2}
\providecommand\natexlab[1]{#1}
\providecommand\showeprint[2][]{arXiv:#2}

\bibitem[\protect\citeauthoryear{Anderson}{Anderson}{2009}]%
        {anderson2009kernel}
\bibfield{author}{\bibinfo{person}{Tessa~K Anderson}.}
  \bibinfo{year}{2009}\natexlab{}.
\newblock \showarticletitle{Kernel density estimation and K-means clustering to
  profile road accident hotspots}.
\newblock \bibinfo{journal}{\emph{Accident Analysis \& Prevention}}
  \bibinfo{volume}{41}, \bibinfo{number}{3} (\bibinfo{year}{2009}),
  \bibinfo{pages}{359--364}.
\newblock


\bibitem[\protect\citeauthoryear{Ben-David, Blitzer, Crammer, Kulesza, Pereira,
  and Vaughan}{Ben-David et~al\mbox{.}}{2010}]%
        {ben2010theory}
\bibfield{author}{\bibinfo{person}{Shai Ben-David}, \bibinfo{person}{John
  Blitzer}, \bibinfo{person}{Koby Crammer}, \bibinfo{person}{Alex Kulesza},
  \bibinfo{person}{Fernando Pereira}, {and} \bibinfo{person}{Jennifer~Wortman
  Vaughan}.} \bibinfo{year}{2010}\natexlab{}.
\newblock \showarticletitle{A theory of learning from different domains}.
\newblock \bibinfo{journal}{\emph{Machine learning}} \bibinfo{volume}{79},
  \bibinfo{number}{1-2} (\bibinfo{year}{2010}), \bibinfo{pages}{151--175}.
\newblock


\bibitem[\protect\citeauthoryear{B{\'\i}l, Andr{\'a}{\v{s}}ik, and
  Jano{\v{s}}ka}{B{\'\i}l et~al\mbox{.}}{2013}]%
        {bil2013identification}
\bibfield{author}{\bibinfo{person}{Michal B{\'\i}l}, \bibinfo{person}{Richard
  Andr{\'a}{\v{s}}ik}, {and} \bibinfo{person}{Zbyn{\v{e}}k Jano{\v{s}}ka}.}
  \bibinfo{year}{2013}\natexlab{}.
\newblock \showarticletitle{Identification of hazardous road locations of
  traffic accidents by means of kernel density estimation and cluster
  significance evaluation}.
\newblock \bibinfo{journal}{\emph{Accident Analysis \& Prevention}}
  \bibinfo{volume}{55} (\bibinfo{year}{2013}), \bibinfo{pages}{265--273}.
\newblock


\bibitem[\protect\citeauthoryear{Black and Villarini}{Black and
  Villarini}{2018}]%
        {black2018effects}
\bibfield{author}{\bibinfo{person}{Alan~W Black} {and}
  \bibinfo{person}{Gabriele Villarini}.} \bibinfo{year}{2018}\natexlab{}.
\newblock \showarticletitle{Effects of methodological decisions on
  rainfall-related crash relative risk estimates}.
\newblock \bibinfo{journal}{\emph{Accident Analysis \& Prevention}}
  (\bibinfo{year}{2018}).
\newblock


\bibitem[\protect\citeauthoryear{Boeing}{Boeing}{2018}]%
        {boeing2018urban}
\bibfield{author}{\bibinfo{person}{Geoff Boeing}.}
  \bibinfo{year}{2018}\natexlab{}.
\newblock \showarticletitle{Urban Spatial Order: Street Network Orientation,
  Configuration, and Entropy}.
\newblock \bibinfo{journal}{\emph{arXiv preprint arXiv:1808.00600}}
  (\bibinfo{year}{2018}).
\newblock


\bibitem[\protect\citeauthoryear{Chen, Song, Yamada, and Shibasaki}{Chen
  et~al\mbox{.}}{2016}]%
        {chen2016learning}
\bibfield{author}{\bibinfo{person}{Quanjun Chen}, \bibinfo{person}{Xuan Song},
  \bibinfo{person}{Harutoshi Yamada}, {and} \bibinfo{person}{Ryosuke
  Shibasaki}.} \bibinfo{year}{2016}\natexlab{}.
\newblock \showarticletitle{Learning Deep Representation from Big and
  Heterogeneous Data for Traffic Accident Inference.}. In
  \bibinfo{booktitle}{\emph{AAAI}}. \bibinfo{pages}{338--344}.
\newblock


\bibitem[\protect\citeauthoryear{Chung, Kim, and Cheon}{Chung
  et~al\mbox{.}}{2018}]%
        {chung2018framework}
\bibfield{author}{\bibinfo{person}{Younshik Chung}, \bibinfo{person}{Seonjung
  Kim}, {and} \bibinfo{person}{Seunghoon Cheon}.}
  \bibinfo{year}{2018}\natexlab{}.
\newblock \showarticletitle{A Framework for Modelling Crash Likelihood
  Information Under Rainy Weather Conditions}. In
  \bibinfo{booktitle}{\emph{International Conference on Applied Human Factors
  and Ergonomics}}. Springer, \bibinfo{pages}{823--832}.
\newblock


\bibitem[\protect\citeauthoryear{Csurka}{Csurka}{2017}]%
        {csurka2017domain}
\bibfield{author}{\bibinfo{person}{Gabriela Csurka}.}
  \bibinfo{year}{2017}\natexlab{}.
\newblock \showarticletitle{Domain adaptation for visual applications: A
  comprehensive survey}.
\newblock \bibinfo{journal}{\emph{arXiv preprint arXiv:1702.05374}}
  (\bibinfo{year}{2017}).
\newblock


\bibitem[\protect\citeauthoryear{Das, Dutta, Jalayer, Bibeka, and Wu}{Das
  et~al\mbox{.}}{2018}]%
        {das2018factors}
\bibfield{author}{\bibinfo{person}{Subasish Das}, \bibinfo{person}{Anandi
  Dutta}, \bibinfo{person}{Mohammad Jalayer}, \bibinfo{person}{Apoorba Bibeka},
  {and} \bibinfo{person}{Lingtao Wu}.} \bibinfo{year}{2018}\natexlab{}.
\newblock \showarticletitle{Factors influencing the patterns of wrong-way
  driving crashes on freeway exit ramps and median crossovers: Exploration
  using ‘Eclat’association rules to promote safety}.
\newblock \bibinfo{journal}{\emph{International Journal of Transportation
  Science and Technology}} \bibinfo{volume}{7}, \bibinfo{number}{2}
  (\bibinfo{year}{2018}), \bibinfo{pages}{114--123}.
\newblock


\bibitem[\protect\citeauthoryear{Deublein, Schubert, Adey, K{\"o}hler, and
  Faber}{Deublein et~al\mbox{.}}{2013}]%
        {deublein2013prediction}
\bibfield{author}{\bibinfo{person}{Markus Deublein}, \bibinfo{person}{Matthias
  Schubert}, \bibinfo{person}{Bryan~T Adey}, \bibinfo{person}{Jochen
  K{\"o}hler}, {and} \bibinfo{person}{Michael~H Faber}.}
  \bibinfo{year}{2013}\natexlab{}.
\newblock \showarticletitle{Prediction of road accidents: A Bayesian
  hierarchical approach}.
\newblock \bibinfo{journal}{\emph{Accident Analysis \& Prevention}}
  \bibinfo{volume}{51} (\bibinfo{year}{2013}), \bibinfo{pages}{274--291}.
\newblock


\bibitem[\protect\citeauthoryear{Dietrich, Gray, McNamara, Poikola, Pollock,
  Tait, Zijlstra, et~al\mbox{.}}{Dietrich et~al\mbox{.}}{2009}]%
        {dietrich2009open}
\bibfield{author}{\bibinfo{person}{Daniel Dietrich}, \bibinfo{person}{Jonathan
  Gray}, \bibinfo{person}{Tim McNamara}, \bibinfo{person}{Antti Poikola},
  \bibinfo{person}{P Pollock}, \bibinfo{person}{Julian Tait},
  \bibinfo{person}{Ton Zijlstra}, {et~al\mbox{.}}}
  \bibinfo{year}{2009}\natexlab{}.
\newblock \showarticletitle{Open data handbook}.
\newblock \bibinfo{journal}{\emph{Open Knowledge International}}
  (\bibinfo{year}{2009}).
\newblock


\bibitem[\protect\citeauthoryear{Fawcett, Thorpe, Matthews, and Kremer}{Fawcett
  et~al\mbox{.}}{2017}]%
        {fawcett2017novel}
\bibfield{author}{\bibinfo{person}{Lee Fawcett}, \bibinfo{person}{Neil Thorpe},
  \bibinfo{person}{Joseph Matthews}, {and} \bibinfo{person}{Karsten Kremer}.}
  \bibinfo{year}{2017}\natexlab{}.
\newblock \showarticletitle{A novel Bayesian hierarchical model for road safety
  hotspot prediction}.
\newblock \bibinfo{journal}{\emph{Accident Analysis \& Prevention}}
  \bibinfo{volume}{99} (\bibinfo{year}{2017}), \bibinfo{pages}{262--271}.
\newblock


\bibitem[\protect\citeauthoryear{Fleiss and Cohen}{Fleiss and Cohen}{1973}]%
        {fleiss1973equivalence}
\bibfield{author}{\bibinfo{person}{Joseph~L Fleiss} {and}
  \bibinfo{person}{Jacob Cohen}.} \bibinfo{year}{1973}\natexlab{}.
\newblock \showarticletitle{The equivalence of weighted kappa and the
  intraclass correlation coefficient as measures of reliability}.
\newblock \bibinfo{journal}{\emph{Educational and psychological measurement}}
  \bibinfo{volume}{33}, \bibinfo{number}{3} (\bibinfo{year}{1973}),
  \bibinfo{pages}{613--619}.
\newblock


\bibitem[\protect\citeauthoryear{Geurts, Thomas, and Wets}{Geurts
  et~al\mbox{.}}{2005}]%
        {geurts2005understanding}
\bibfield{author}{\bibinfo{person}{Karolien Geurts}, \bibinfo{person}{Isabelle
  Thomas}, {and} \bibinfo{person}{Geert Wets}.}
  \bibinfo{year}{2005}\natexlab{}.
\newblock \showarticletitle{Understanding spatial concentrations of road
  accidents using frequent item sets}.
\newblock \bibinfo{journal}{\emph{Accident Analysis \& Prevention}}
  \bibinfo{volume}{37}, \bibinfo{number}{4} (\bibinfo{year}{2005}),
  \bibinfo{pages}{787--799}.
\newblock


\bibitem[\protect\citeauthoryear{Geurts, Wets, Brijs, and Vanhoof}{Geurts
  et~al\mbox{.}}{2003}]%
        {geurts2003profiling}
\bibfield{author}{\bibinfo{person}{Karolien Geurts}, \bibinfo{person}{Geert
  Wets}, \bibinfo{person}{Tom Brijs}, {and} \bibinfo{person}{Koen Vanhoof}.}
  \bibinfo{year}{2003}\natexlab{}.
\newblock \showarticletitle{Profiling of high-frequency accident locations by
  use of association rules}.
\newblock \bibinfo{journal}{\emph{Transportation Research Record: Journal of
  the Transportation Research Board}} \bibinfo{number}{1840}
  (\bibinfo{year}{2003}), \bibinfo{pages}{123--130}.
\newblock


\bibitem[\protect\citeauthoryear{Gomes}{Gomes}{2013}]%
        {gomes2013influence}
\bibfield{author}{\bibinfo{person}{Sandra~Vieira Gomes}.}
  \bibinfo{year}{2013}\natexlab{}.
\newblock \showarticletitle{The influence of the infrastructure characteristics
  in urban road accidents occurrence}.
\newblock \bibinfo{journal}{\emph{Accident Analysis \& Prevention}}
  \bibinfo{volume}{60} (\bibinfo{year}{2013}), \bibinfo{pages}{289--297}.
\newblock


\bibitem[\protect\citeauthoryear{He, Zhang, Ren, and Sun}{He
  et~al\mbox{.}}{2016}]%
        {he2016deep}
\bibfield{author}{\bibinfo{person}{Kaiming He}, \bibinfo{person}{Xiangyu
  Zhang}, \bibinfo{person}{Shaoqing Ren}, {and} \bibinfo{person}{Jian Sun}.}
  \bibinfo{year}{2016}\natexlab{}.
\newblock \showarticletitle{Deep residual learning for image recognition}. In
  \bibinfo{booktitle}{\emph{Proceedings of the IEEE conference on computer
  vision and pattern recognition}}. \bibinfo{pages}{770--778}.
\newblock


\bibitem[\protect\citeauthoryear{Long, Cao, Wang, and Jordan}{Long
  et~al\mbox{.}}{2015}]%
        {long2015learning}
\bibfield{author}{\bibinfo{person}{Mingsheng Long}, \bibinfo{person}{Yue Cao},
  \bibinfo{person}{Jianmin Wang}, {and} \bibinfo{person}{Michael~I Jordan}.}
  \bibinfo{year}{2015}\natexlab{}.
\newblock \showarticletitle{Learning transferable features with deep adaptation
  networks}.
\newblock \bibinfo{journal}{\emph{arXiv preprint arXiv:1502.02791}}
  (\bibinfo{year}{2015}).
\newblock


\bibitem[\protect\citeauthoryear{Najjar, Kaneko, and Miyanaga}{Najjar
  et~al\mbox{.}}{2017}]%
        {najjar2017combining}
\bibfield{author}{\bibinfo{person}{Alameen Najjar}, \bibinfo{person}{Shun'ichi
  Kaneko}, {and} \bibinfo{person}{Yoshikazu Miyanaga}.}
  \bibinfo{year}{2017}\natexlab{}.
\newblock \showarticletitle{Combining Satellite Imagery and Open Data to Map
  Road Safety.}. In \bibinfo{booktitle}{\emph{AAAI}}.
  \bibinfo{pages}{4524--4530}.
\newblock


\bibitem[\protect\citeauthoryear{Nieminen, Lehtonen, and Linna}{Nieminen
  et~al\mbox{.}}{2002}]%
        {nieminen2002population}
\bibfield{author}{\bibinfo{person}{Seppo Nieminen}, \bibinfo{person}{Olli-Pekka
  Lehtonen}, {and} \bibinfo{person}{Miika Linna}.}
  \bibinfo{year}{2002}\natexlab{}.
\newblock \showarticletitle{Population density and occurrence of accidents in
  finland}.
\newblock \bibinfo{journal}{\emph{Prehospital and disaster medicine}}
  \bibinfo{volume}{17}, \bibinfo{number}{4} (\bibinfo{year}{2002}),
  \bibinfo{pages}{206--208}.
\newblock


\bibitem[\protect\citeauthoryear{Organization}{Organization}{2015}]%
        {world2015global}
\bibfield{author}{\bibinfo{person}{World~Health Organization}.}
  \bibinfo{year}{2015}\natexlab{}.
\newblock \bibinfo{booktitle}{\emph{Global status report on road safety 2015}}.
\newblock \bibinfo{publisher}{World Health Organization}.
\newblock


\bibitem[\protect\citeauthoryear{Pulugurtha, Duddu, and Kotagiri}{Pulugurtha
  et~al\mbox{.}}{2013}]%
        {pulugurtha2013traffic}
\bibfield{author}{\bibinfo{person}{Srinivas~S Pulugurtha},
  \bibinfo{person}{Venkata~Ramana Duddu}, {and} \bibinfo{person}{Yashaswi
  Kotagiri}.} \bibinfo{year}{2013}\natexlab{}.
\newblock \showarticletitle{Traffic analysis zone level crash estimation models
  based on land use characteristics}.
\newblock \bibinfo{journal}{\emph{Accident Analysis \& Prevention}}
  \bibinfo{volume}{50} (\bibinfo{year}{2013}), \bibinfo{pages}{678--687}.
\newblock


\bibitem[\protect\citeauthoryear{Schmucker, Seifert, Stengel, Matthes,
  Ottersbach, and Ekkernkamp}{Schmucker et~al\mbox{.}}{2010}]%
        {schmucker2010road}
\bibfield{author}{\bibinfo{person}{Uli Schmucker}, \bibinfo{person}{J Seifert},
  \bibinfo{person}{Dirk Stengel}, \bibinfo{person}{Gerrit Matthes},
  \bibinfo{person}{Caspar Ottersbach}, {and} \bibinfo{person}{Axel
  Ekkernkamp}.} \bibinfo{year}{2010}\natexlab{}.
\newblock \showarticletitle{Road traffic crashes in developing countries}.
\newblock \bibinfo{journal}{\emph{Der Unfallchirurg}} \bibinfo{volume}{113},
  \bibinfo{number}{5} (\bibinfo{year}{2010}), \bibinfo{pages}{373--377}.
\newblock


\bibitem[\protect\citeauthoryear{Sun and Saenko}{Sun and Saenko}{2016}]%
        {sun2016deep}
\bibfield{author}{\bibinfo{person}{Baochen Sun} {and} \bibinfo{person}{Kate
  Saenko}.} \bibinfo{year}{2016}\natexlab{}.
\newblock \showarticletitle{Deep coral: Correlation alignment for deep domain
  adaptation}. In \bibinfo{booktitle}{\emph{European Conference on Computer
  Vision}}. Springer, \bibinfo{pages}{443--450}.
\newblock


\bibitem[\protect\citeauthoryear{Tamerius, Zhou, Mantilla, and
  Greenfield-Huitt}{Tamerius et~al\mbox{.}}{2016}]%
        {tamerius2016precipitation}
\bibfield{author}{\bibinfo{person}{JD Tamerius}, \bibinfo{person}{X Zhou},
  \bibinfo{person}{R Mantilla}, {and} \bibinfo{person}{T Greenfield-Huitt}.}
  \bibinfo{year}{2016}\natexlab{}.
\newblock \showarticletitle{Precipitation Effects on Motor Vehicle Crashes Vary
  by Space, Time, and Environmental Conditions}.
\newblock \bibinfo{journal}{\emph{Weather, Climate, and Society}}
  \bibinfo{volume}{8}, \bibinfo{number}{4} (\bibinfo{year}{2016}),
  \bibinfo{pages}{399--407}.
\newblock


\bibitem[\protect\citeauthoryear{Tzeng, Hoffman, Zhang, Saenko, and
  Darrell}{Tzeng et~al\mbox{.}}{2014}]%
        {tzeng2014deep}
\bibfield{author}{\bibinfo{person}{Eric Tzeng}, \bibinfo{person}{Judy Hoffman},
  \bibinfo{person}{Ning Zhang}, \bibinfo{person}{Kate Saenko}, {and}
  \bibinfo{person}{Trevor Darrell}.} \bibinfo{year}{2014}\natexlab{}.
\newblock \showarticletitle{Deep domain confusion: Maximizing for domain
  invariance}.
\newblock \bibinfo{journal}{\emph{arXiv preprint arXiv:1412.3474}}
  (\bibinfo{year}{2014}).
\newblock


\bibitem[\protect\citeauthoryear{Wu, Han, Lin, Uzunbas, Goldstein, Lim, and
  Davis}{Wu et~al\mbox{.}}{2018}]%
        {wu2018dcan}
\bibfield{author}{\bibinfo{person}{Zuxuan Wu}, \bibinfo{person}{Xintong Han},
  \bibinfo{person}{Yen-Liang Lin}, \bibinfo{person}{Mustafa~Gkhan Uzunbas},
  \bibinfo{person}{Tom Goldstein}, \bibinfo{person}{Ser~Nam Lim}, {and}
  \bibinfo{person}{Larry~S Davis}.} \bibinfo{year}{2018}\natexlab{}.
\newblock \showarticletitle{DCAN: Dual Channel-wise Alignment Networks for
  Unsupervised Scene Adaptation}.
\newblock \bibinfo{journal}{\emph{arXiv preprint arXiv:1804.05827}}
  (\bibinfo{year}{2018}).
\newblock


\bibitem[\protect\citeauthoryear{Yuan, Zhou, and Yang}{Yuan
  et~al\mbox{.}}{2018}]%
        {yuan2018hetero}
\bibfield{author}{\bibinfo{person}{Zhuoning Yuan}, \bibinfo{person}{Xun Zhou},
  {and} \bibinfo{person}{Tianbao Yang}.} \bibinfo{year}{2018}\natexlab{}.
\newblock \showarticletitle{Hetero-ConvLSTM: A Deep Learning Approach to
  Traffic Accident Prediction on Heterogeneous Spatio-Temporal Data}. In
  \bibinfo{booktitle}{\emph{Proceedings of the 24th ACM SIGKDD International
  Conference on Knowledge Discovery \& Data Mining}}. ACM,
  \bibinfo{pages}{984--992}.
\newblock


\bibitem[\protect\citeauthoryear{Zhou, Khosla, Lapedriza, Oliva, and
  Torralba}{Zhou et~al\mbox{.}}{2016}]%
        {zhou2016learning}
\bibfield{author}{\bibinfo{person}{Bolei Zhou}, \bibinfo{person}{Aditya
  Khosla}, \bibinfo{person}{Agata Lapedriza}, \bibinfo{person}{Aude Oliva},
  {and} \bibinfo{person}{Antonio Torralba}.} \bibinfo{year}{2016}\natexlab{}.
\newblock \showarticletitle{Learning deep features for discriminative
  localization}. In \bibinfo{booktitle}{\emph{Proceedings of the IEEE
  Conference on Computer Vision and Pattern Recognition}}.
  \bibinfo{pages}{2921--2929}.
\newblock


\bibitem[\protect\citeauthoryear{Zhu, Kong, Zheng, Fu, and Tian}{Zhu
  et~al\mbox{.}}{2017}]%
        {zhu2017part}
\bibfield{author}{\bibinfo{person}{Fuqing Zhu}, \bibinfo{person}{Xiangwei
  Kong}, \bibinfo{person}{Liang Zheng}, \bibinfo{person}{Haiyan Fu}, {and}
  \bibinfo{person}{Qi Tian}.} \bibinfo{year}{2017}\natexlab{}.
\newblock \showarticletitle{Part-based deep hashing for large-scale person
  re-identification}.
\newblock \bibinfo{journal}{\emph{IEEE Transactions on Image Processing}}
  \bibinfo{volume}{26}, \bibinfo{number}{10} (\bibinfo{year}{2017}),
  \bibinfo{pages}{4806--4817}.
\newblock


\end{thebibliography}
